\documentclass[10 pt, conference]{ieeeconf}  
\IEEEoverridecommandlockouts                              
\usepackage{graphicx} 
\usepackage{amssymb,mathtools,mathrsfs}  
\usepackage{algorithm, algorithmic}
\usepackage[hidelinks]{hyperref}
\usepackage{parskip}
\usepackage{xcolor}

\newtheorem{definition}{Definition}

\newcommand{\rev}[1]{\textcolor{black}{#1}}

\title{\LARGE \bf
Real-Time Multi-Contact Model Predictive Control via ADMM
}

\author{Alp Aydinoglu$^{1}$ and Michael Posa$^{1}$
\thanks{*This work was supported by the National Science Foundation under Grants No. CMMI-1830218 and EFRI-1935294.
}
    \thanks{$^{1}$Alp Aydinoglu and Michael Posa are with the General Robotics, Automation, Sensing and Perception (GRASP) Laboratory, University of Pennsylvania, Philadelphia, PA 19104, USA.
    {\tt\small \{alpayd, posa\}@seas.upenn.edu}}%
}

\begin{document}
\maketitle
\thispagestyle{empty}
\pagestyle{empty}


\begin{abstract}
We propose a hybrid model predictive control algorithm, consensus complementarity control (C3), for systems that make and break contact with their environment.
Many state-of-the-art controllers for tasks which require initiating contact with the environment, such as locomotion and manipulation, require \emph{a priori} mode schedules or are so computationally complex that they cannot run at real-time rates.
We present a method, based on the alternating direction method of multipliers (ADMM), capable of high-speed reasoning over potential contact events.
Via a consensus formulation, our approach enables parallelization of the contact scheduling problem.
We validate our results on three numerical examples, including two frictional contact problems, and physical experimentation on an underactuated multi-contact system.
\end{abstract}

\section{Introduction}
For many important tasks such as manipulation and locomotion, robots need to make and break contact with the environment. 
Even though such multi-contact systems are common, they are notoriously hard to control. 
The main challenge is finding policies and/or trajectories that explicitly consider the interaction of the robot with its environment in order to enable stable, robust motion.
For a wide range of problems, it is computationally challenging to discover control policies and/or trajectories \cite{posa2014direct} and the methods are not suitable for running in real-time speeds for complex problems.

Model predictive control (MPC) is one of the most powerful tools for automatic control \cite{garcia1989model, borrelli2017predictive}. 
Predominant in robotics are MPC-based methods utilizing linearization, leading to quadratic programs which can be solved efficiently \cite{ding2019real, zhakatayev2017successive}. However, for multi-contact systems, the algorithm must also decide when to initiate or break contact. As a result, linearizations are no longer appropriate and a hybrid formulation is required. 

When linearizations are not viable, the resulting MPC algorithm includes the hybrid elements that result from making and breaking of contact. Due to the computational complexity of hybrid MPC, these methods can use a predefined sequence of contacts \cite{mastalli2020crocoddyl, sleiman2021unified}, specialize for legged systems \cite{winkler2018gait, bledt2020regularized, lee2020learning, kuindersma2016optimization}, quasi-static manipulation \cite{hogan2020reactive}, rely on an offline phase \cite{marcucci2017approximate, aydinoglu2021stability}, or approximate the dynamics via smooth contact models and change the dynamics (diagonal approximations of contact matrix) to enable high performance \cite{tassa2012synthesis, koenemann2015whole, kumar2014real}.
An unifying theme across much of prior work is that these approaches require significant domain knowledge for simplifying the search for appropriate mode sequences.
On the other hand, developing MPC based control techniques that can reason about contact events and that do not require domain knowledge is a much harder task. Marcucci and Tedrake \cite{marcucci2020warm} utilize mixed-integer programming to optimize over control actions and mode sequences, but the method requires significant online computation time. Recently, contact-implicit control with a primal-dual interior-point method has been explored with impressive results on simulation but few details are available and the method has not yet been validated on a physical system \cite{cleac2021linear}.
Operator splitting has been used for solving hybrid MPC problems, with local guarantees, but this approach is not suitable for multi-contact control due to exponential increase in the number of constraints based on number of contacts \cite{frick2019low}. 

In contrast to prior work, here we focus on a specific structure in multi-contact robotics and exploit this structure via ADMM \cite{boyd2011distributed}. It has been explored for general mixed integer programs \cite{stellato2018embedded,takapoui2020simple, park2017general} but has not been explored for multi-contact systems. In this work, we will focus on local linear complementarity system \cite{heemels2000linear} approximations of the nonlinear plants and explore ADMM for this particular class of multi-contact systems.

The primary contribution of this paper is an algorithm, consensus complementarity control (C3), for solving the hybrid MPC problem approximately for multi-contact systems. We exploit the distributed nature of ADMM and demonstrate that the hard part of the problem, reasoning about contact events, can be parallelized. This enables our algorithm to be fast ($\sim 16$ Hz for a system with ten states and three contacts), robust to disturbances and also minimizes the effect of control horizon on the run-time of the algorithm. We also demonstrate the effectiveness of our method on numerical examples and present results on a hardware setup.
Experimental validation of contact-implicit MPC algorithms is extremely rare. To the best of our knowledge, these are the first real-time control results on an underactuated system as complex and dynamic as the hybrid cart-pole.

\section{Background}

The function $\mathcal{I}_\mathcal{A}$ is the $0-\infty$ indicator function for an arbitrary set $\mathcal{A}$ such that $\mathcal{I}_\mathcal{A}(z) = 0$ if $z \in \mathcal{A}$ and $\mathcal{I}_\mathcal{A}(z) = \infty$ if $z \notin \mathcal{A}$. $\mathbb{N}_0$ denotes natural numbers with zero.

\subsection{Linear Complementarity Problem}

In this work, we utilize complementarity problems to represent the contact forces. These models are common in modeling multi-contact robotics problems \cite{stewart2000implicit} and can also be efficiently learned from data \cite{pfrommer2020contactnets}.
The theory of linear complementarity problems (LCP) is well established \cite{cottle2009linear}.
\begin{definition} \label{def: LCS_definition}
	Given a vector $q \in \mathbb{R}^m$, and a matrix ${F \in \mathbb{R}^{m \times m}}$, the $\text{LCP}(q,F)$  describes the following mathematical program:
	\begin{alignat}{2}
	\label{LCS_definiton}
	\notag & \underset{}{\text{find}} && \lambda \in \mathbb{R}^m \\
	\notag & \text{subject to}  \quad && y = F \lambda + q,\\
	\notag & \quad && 0 \leq \lambda \perp y \geq 0.
	\end{alignat}
\end{definition}
Here, the vector $\lambda$ typically represents the contact forces and slack variables (\cite{stewart2000implicit}),  $y$ represents the gap function and orthogonality constraint embeds the hybrid structure.

\subsection{Linear Complementarity Systems}
Linear complementarity systems (LCS) are useful because they formulate the discrete-time dynamics of a multi-contact system. We define an LCS as a difference equation coupled with a variable that is the solution of an LCP.

\begin{definition}
	A linear complementarity system describes the  trajectories $( x_k )_{k \in \mathbb{N}_{0}}$ and $( \lambda_k )_{k \in \mathbb{N}_{0}}$ for an input sequence $( u_k )_{k \in \mathbb{N}_{0}}$ starting from $x_0$ such that
	\begin{equation}
	\label{eq:LCS}
	\begin{aligned}
	& x_{k+1} = A x_k + B u_k + D\lambda_k + d,\\
	& 0 \leq \lambda_k \perp Ex_k +  F \lambda_k + H u_k + c \geq 0,
	\end{aligned}
	\end{equation}
	where $x_k \in \mathbb{R}^{n_x}$, $\lambda_k \in \mathbb{R}^{n_{\lambda}}$, $u_k \in \mathbb{R}^{n_u}$, $A \in \mathbb{R}^{n_x \times n_x}$, $B \in \mathbb{R}^{n_x \times n_u}$, $D \in \mathbb{R}^{n_x \times n_\lambda}$, $d \in \mathbb{R}^{n_x}$, $E \in \mathbb{R}^{n_{\lambda} \times n_x}$, $F \in \mathbb{R}^{n_x \times n_\lambda}$, $H \in \mathbb{R}^{n_\lambda \times n_u}$ and $c \in \mathbb{R}^{n_\lambda}$.
\end{definition}

For a given $k$, $x_k$ and $u_k$, the corresponding complementarity variable $\lambda_k$ can be found by solving $\text{LCP}(E x_k + H u_k + c, F)$ (see Definition \ref{def: LCS_definition}). Similarly, $x_{k+1}$ can be computed using the first equation in \eqref{eq:LCS} when $x_k, u_k$ and $\lambda_k$ are known. \rev{We focus on cases where $( x_k )_{k \in \mathbb{N}_{0}}$ is unique, and note that this holds for the examples in the paper. While non-uniqueness can arise due to rigidity \cite{halm2019modeling}, control in these pathological settings is outside the scope of this paper.}

\section{Model Predictive Control of Multi-contact Systems}

\label{sec:ADMM_main}

\subsection{Problem Formulation}

In this work, we want to solve the following mathematical optimization problem:
\begin{equation}
\label{eq:MPC_original}
\begin{aligned}
\min_{x_k, \lambda_k, u_k} \quad & \sum_{k=0}^{N-1} (x_k^T Q_k x_k + u_k^T R_k u_k)  + x_N^T Q_N x_N \\
\textrm{s.t.} \quad &x_{k+1} = A x_k + B u_k + D \lambda_k + d, \\
& E x_k + F \lambda_k + H u_k + c \geq 0, \\
& \lambda_k \geq 0, \\
& \lambda_k^T (E x_k + F \lambda_k + H u_k + c) = 0, \\
& (\boldsymbol x, \boldsymbol \lambda, \boldsymbol u) \in \mathcal{C}, \\
& \text{for} \; k = 0, \ldots, N-1, \text{given} \; x_0,
\end{aligned}
\end{equation}
where $N$ is the planning horizon, $Q_k, Q_N$ are positive semidefinite matrices, $R_k$ are positive definite matrices and $\mathcal{C}$ is a convex set (e.g. input bounds, safety constraints, or goal conditions) and $\boldsymbol{x}^T = [x_1^T, \ldots, x_{N}^T]$, $\boldsymbol{\lambda}^T = [\lambda_0^T, \lambda_1^T, \ldots, \lambda_{N-1}^T]$, $\boldsymbol{u}^T = [u_0^T, u_1^T, \ldots, u_{N-1}^T]$. 

Given $x_0$, one solves the optimization \eqref{eq:MPC_original} and applies $u_0$ to the plant and repeats the process in every time step in a receding horizon manner.

\subsection{Mixed Integer Formulation}

One straightforward, but computationally expensive, approach to solving \eqref{eq:MPC_original} is via a mixed integer formulation which exchanges the non-convex orthogonality constraints for binary variables:
\begin{equation}
\label{eq:MPC_MIQP}
\begin{aligned}
\min_{x_k, \lambda_k, u_k, s_k} \quad & \sum_{k=0}^{N-1} (x_k^T Q_k x_k + u_k^T R_k u_k)  + x_N^T Q_N x_N \\
\textrm{s.t.} \quad &x_{k+1} = A x_k + B u_k + D \lambda_k + d, \\
& M s_k \geq E x_k + F \lambda_k + H u_k + c \geq 0, \\
& M (\mathbf{1} - s_k) \geq \lambda_k \geq 0, \\
& (\boldsymbol x, \boldsymbol \lambda, \boldsymbol u) \in \mathcal{C}, s_k \in \{ 0,1 \}^{n_\lambda}, \\
& \text{for} \; k = 0, \ldots, N-1, \text{given} \; x_0,
\end{aligned}
\end{equation}
where $\mathbf{1}$ is a vector of ones, $M$ is a scalar used for the big M method and $s_k$ are the binary variables. 
This approach requires solving $2^{N n_\lambda}$ quadratic programs as a worst case.
This method is popular because, in practice, it is much faster than this worst case analysis. Even still, for the multi-contact problems explored in this work, directly solving \eqref{eq:MPC_MIQP} via state-of-the-art commercial solvers remains too slow for real-time control. Methods that learn the MIQP problem offline are promising, but this line of work requires large-scale training on every problem instance \cite{cauligi2020learning}, \cite{aydinoglu2021stability}.


\subsection{Consensus Complementarity Control (C3)}

Utilizing ADMM, we will solve \eqref{eq:MPC_original} more quickly than with mixed integer formulations. 
Since the problem we are addressing is non-convex, our method is not guaranteed to find the global solution or converge unlike \rev{MIQP-based} approaches, but is significantly faster.
We rewrite \eqref{eq:MPC_original}, equivalently, in the consensus form \cite{park2017general} where we create copies (named $\delta_k$) of variables $z_k^T = [x_k^T, \lambda_k^T, u_k^T]$ and move the constraints into the objective function using $0 -\infty$ indicator functions:
\begin{equation}
\label{eq:MPC_consensus}
\begin{aligned}
\min_{ z } \quad & c(z) + \mathcal{I}_\mathcal{D} ( z  ) + \mathcal{I}_\mathcal{C} ( z  ) + \sum_{k=0}^{N-1} \mathcal{I}_{\mathcal{H}_k} (\delta_k) \\
\textrm{s.t.} \quad &z_k = \delta_k, \; \forall k,
\end{aligned}
\end{equation}
where $z^T = [z_0^T, z_1^T, \ldots, z_{N}^T]$, $c(z)$ is the cost function in \eqref{eq:MPC_original}, the set $\mathcal{D}$ includes the dynamics constraints:
\begin{equation*}
	\cap_{k=0}^{N-1} \{ z : x_{k+1} = A x_k + B u_k + D \lambda_k + d  \},
\end{equation*}
the sets $\mathcal{H}_k$ represent the LCP (contact) constraints:
\begin{align*}
	\mathcal{H}_k = \{&(x_k, \lambda_k, u_k) : E x_k + F \lambda_k + H u_k + c \geq 0, \\
	& \lambda_k \geq 0, \lambda_k^T (E x_k + F \lambda_k + H u_k + c) = 0  \}.
\end{align*}
Note that we leverage the time-dependent structure in the complementarity constraints to separate $\delta_k$'s.

The general augmented Lagrangian (\cite{boyd2011distributed}, Section 3.4.2.) for the problem in consensus form \eqref{eq:MPC_consensus} is:
\begin{align*}
	\mathcal{L} & (z,  \delta, w) = c(z) + \mathcal{I}_\mathcal{D} ( z ) + \mathcal{I}_\mathcal{C} ( z )   \\
	&+ \sum_{i=0}^{N-1} \big( \mathcal{I}_{\mathcal{H}_k} (\delta_k) + ( r_k^T G_k r_k - w_k^T G_k w_k) \big ),
\end{align*}
where $w^T = [w_0^T, w_1^T, \ldots, w_{N-1}^T]$, $w_k$ are scaled dual variables, $r_k = z_k - \delta_k + w_k$, $G_k$ is a positive definite matrix, and $\delta^T = [\delta_0^T, \delta_1^T, \ldots, \delta_{N-1}^T]$. Observe that the standard augmented Lagrangian \cite{park2017general} is recovered for $G_k = \rho_k I$.

In order to solve \eqref{eq:MPC_consensus}, we apply the classical ADMM algorithm, consisting of the following operations:
\begin{align}
	\label{eq:bir}
	& z^{i+1} = \text{argmin}_z \mathcal{L} (z, \delta_0^i, \ldots, \delta_{N-1}^i, w_1^i, \ldots w_{N-1}^i ), \\
	\label{eq:iki}
	& \delta_k^{i+1} = \text{argmin}_{\delta_k}  \mathcal{L} (z^{i+1}, \ldots, \delta_k, \ldots, \delta_{N-1}^i, w_0^i, \ldots w_{N-1}^i ), \\
	\label{eq:uc}
	& w_k^{i+1} = w_k^i + z^{i+1}_k - \delta_k^{i+1}.
\end{align}
Here, \eqref{eq:bir} requires solving a quadratic program, \eqref{eq:iki} is a projection onto the LCP constraints and \eqref{eq:uc} is a dual variable update. Next, we analyze these operations in the given order.

\subsubsection{Quadratic Step}
Equation \eqref{eq:bir} can be represented by the convex quadratic program
\begin{equation}
\label{eq:quadratic_program}
\begin{aligned}
\min_{z} \quad & c(z) + \sum_{i=0}^{N-1} (z_k - \delta_k^i + w_k^i)^T G_k (z_k - \delta_k^i + w_k^i)\\
\textrm{s.t.} \quad & z \in \mathcal{D} \cap \mathcal{C}.
\end{aligned}
\end{equation}
The linear dynamics constraints are captured by the set $\mathcal{D}$, and the convex inequality constraints on states, inputs, contact forces are captured by $\mathcal{C}$. 
The complementarity constraints do not explicitly appear, but their influence is found, iteratively, through the variables $\delta_k^i$.

The QP in \eqref{eq:quadratic_program} can be solved quickly via off-the-shelf solvers and is analogous to solving the MPC problem for a linear system without contact.

\subsubsection{Projection Step}

This step requires projecting onto the LCP constraints $\mathcal{H}_k$ and is the most challenging part of the problem. \eqref{eq:iki} can be represented with the following quadratic program with a non-convex constraint:
\begin{equation}
\label{eq:projection_program}
\begin{aligned}
\min_{\delta_k} \quad & (\delta_k - (z_k^{i+1} + w^i_k) )^T G_k (\delta_k - (z_k^{i+1} + w^i_k) )  \\
\textrm{s.t.} \quad & \delta_k \in \mathcal{H}_k,
\end{aligned}
\end{equation}
where $\delta_k = [(\delta_k^x)^T, (\delta_k^\lambda)^T, (\delta_k^u)^T]$.
In our setting, we will consider three different projections. Two are approximate projections, common for minimization problems over non-convex sets \cite{diamond2018general}.
\paragraph{MIQP Projection}
The projection can be calculated by exactly formulating \eqref{eq:projection_program} as a small-scale MIQP
\begin{equation}
\label{eq:approx_proj}
\begin{aligned}
\min_{\delta_k} \quad & (\delta_k - (z_k^{i+1} + w^i_k) )^T U (\delta_k - (z_k^{i+1} + w^i_k) )  \\
\textrm{s.t.} \quad & E \delta_k^x + F \delta_k^\lambda + H \delta_k^u + c \geq 0, \\
& \delta_k^\lambda \geq 0, \\
& (\delta_k^\lambda)^T ( E \delta_k^x + F \delta_k^\lambda + H \delta_k^u + c ) = 0,
\end{aligned}
\end{equation}
where $U$ is a positive semi-definite matrix. This is a non-convex QP due to the last (orthogonality) constraint.
For $U = G_k$, one recovers the problem in \eqref{eq:projection_program};
however, in our experience, we found significantly improved performance using alternate, but fixed, choices for $U$.
Observe that while \eqref{eq:approx_proj} is non-convex, it is written only in terms of variables corresponding to a single time step $k$. While the original MIQP formulation in \eqref{eq:MPC_MIQP} has $Nn_\lambda$ binary variables, here we have $N$ independent problems, each with $n_\lambda$ binary variables. This decoupling leads to dramatically improved performance (\rev{worst-case $N 2^{N}$ vs $2^{Nn_\lambda}$}).

\paragraph{LCP Projection}
In cases where \eqref{eq:approx_proj} cannot be solved quickly enough, we propose two approximate solutions with faster run-time.
Consider the limiting case where $U$ has no penalty on the force elements. Here, \eqref{eq:approx_proj} can be solved with optimal objective value of $0$ by setting $\delta_k^x = z_k^{(i+1),x} + w_k^{(i),x}$ and $\delta_k^u = z_k^{(i+1),u} + w_k^{(i),u}$. Then, $\delta_k^\lambda$ can be found by solving $\text{LCP}(E \delta_k^x + H \delta_k^u + c,F)$.
We note that this projection is different than shooting based methods since we are simulating the $z_k^{(i+1),x,u} + w_k^{(i),x,u}$ instead of $z_k^{(i+1),x,u}$.


\paragraph{ADMM Projection}
We note that \eqref{eq:approx_proj} is a quadratically constrained quadratic program, where prior work has solved problems of this form using ADMM \cite{park2017general}. 
This leads to nested ADMM algorithms, but this formulation can  produce faster solutions than MIQP solvers without guarantees that it produces a feasible or optimal value. 
Note that this formulation fared poorly when applied directly to the original problem \eqref{eq:MPC_original}, rarely satisfying the complementarity constraints.

Solving the full MIQP \eqref{eq:approx_proj} typically is the best at exploring a wide range of modes, while the LCP projection is usually fastest (depending on matrix $F$). ADMM projection can be viewed as a middle ground.



\begin{algorithm}[t!]
	\caption{Consensus Complementarity Control (C3)}
	\begin{algorithmic}[1]
		\REQUIRE $\delta^0_k, w^0_k, G_k, D, \theta, \rho_k, N, x_0 $	
		\\ \textit{Initialization} : $i=1$
		\WHILE {$i \leq \theta $}
		\STATE Compute $z^{i+1}$ via \eqref{eq:quadratic_program}
		\STATE Compute $\delta_k^{i+1}$ via \eqref{eq:approx_proj}, $\forall k$
		\STATE  $w_k^{i+1} \gets w_k^i + z^{i+1}_k - \delta_k^{i+1}, \forall k$
		\STATE $G_k \gets \rho_k G_k, \forall k$
		\STATE $w_k \gets w_k / \rho_k, \forall k $
		\ENDWHILE
		\RETURN $u_0$
	\end{algorithmic} 
	\label{algortihm_ADMM}
\end{algorithm}

After discussing each of the individual steps, we present the full C3 algorithm (Algorithm \ref{algortihm_ADMM}). Here, both $\delta_k^0$ and $w_k^0$ are usually initialized as zero vectors. $G_k$ are positive definite matrices, $\theta > 0 $ is the number of ADMM steps, $\rho_k > 0$ is the penalty parameter. Given an horizon $N$, and an initial state $x_0$, the algorithm returns $u_0$ which is applied to the system. This procedure is repeated in every time step as in receding horizon controllers. \rev{In the interest of real-time rates, we run a fixed number of ADMM steps ($\theta$).}

\section{Numerical Examples}

For these results, OSQP \cite{osqp} is used to solve quadratic programs and Gurobi \cite{Gurobi} is used for mixed integer programs. PATH \cite{dirkse1995path} and Lemke's algorihm have been used to solve LCP's. SI units (meter, kilogram, second) are used. For all the examples,  $G_k = \rho_k G_{k-1}$ is used where $G_0 = G$ is positive definite.
The experiments are done on a desktop computer with the processor Intel \emph{i7-11800H} and \emph{16GB RAM}. 
Reported run-times include all steps in the algorithm. Aside from these third-party solvers, code was implemented in Python. We expect that further improvements in runtime would be achievable using C++.
The code for all examples is available\footnote{ \url{https://github.com/AlpAydinoglu/coptimal}}. Experiments are also shown in the supplementary video.

\subsection{Cart-pole with Soft Walls}

\label{subsec:cartpole_sim}


We consider a cart-pole that can interact with soft walls as in \cite{marcucci2020warm, aydinoglu2020stabilization}. This is a benchmark in contact-based control algorithms. Here, $x^{(1)}$ represents the position of the cart, $x^{(2)}$ represents the position of the pole and $x^{(3)}, x^{(4)}$ represent their velocities respectively. The forces that affect the pole are described by $\lambda^{(1)}$ and $\lambda^{(2)}$ for right and left walls respectively.

\begin{table}[t]
	\centering
	\caption{Projection Run-time and averaged cost-to-go}
	\label{tab:table_speed}
	\renewcommand{\arraystretch}{1.2}
	\begin{tabular}{|c|c|c|c|c|c|}
		\hline
		\textbf{Projection Method}       & \textbf{Mean} $\pm$ \textbf{Std} ($s$)  & \textbf{Cost} \\ \hline
		\textbf{LCP}   & $ 1.4 \cdot 10^{-5}$ $\pm$ $ 1.8 \cdot 10^{-6}$  & $19.16 $      \\ \hline
		\textbf{MIQP}   & $ 1 \cdot 10^{-3}$ $\pm$ $ 1.2 \cdot 10^{-4}$  & $27.28 $     \\ \hline
		\textbf{ADMM}   & $ 5.3 \cdot 10^{-4}$ $\pm$ $3.3 \cdot 10^{-5}$   & $ 33.58 $     \\ \hline
	\end{tabular}
\end{table}

The model is linearized around $x^{(2)} = 0$ where $m_c=0.978$ is the mass of the cart, $m_p = 0.411$ is the combined mass of the pole and the rod, $l_p = 0.6$ is the length of the pole, $l_c = 0.4267$ is the length of the center of mass position, $k_1 = k_2 = 50$ are the stiffness parameter of the walls, $d=0.35$ is the distance between the origin and the soft walls. Then, we discretize the dynamics using the explicit Euler method with time step $T_s = 0.01$ to obtain the system matrices and use the model in \cite{aydinoglu2020stabilization}. We note that the MIQP formulation (as in \eqref{eq:MPC_MIQP}) runs at approximately $10$ Hz.

We design a controller where $\theta = 10$, $G = 0.1 I$, $\rho_k = 2$, and $N=10$. There is a clear trade-off between solve time and planning horizon \cite{li2021model}.
We test all three projection methods described in Section \ref{sec:ADMM_main} and report run-times (single solve of \eqref{eq:approx_proj}) on Table \ref{tab:table_speed} averaged for 1000 solves.
We also report the average of cost-to-go value assuming all of the methods can run at $100$ Hz (only the MIQP projection cannot). Even though the MIQP projection is usually better at exploring new contacts, it does not always result in a better cost. We note that the controller can run slightly faster than $240$ Hz if LCP-based projection is used. For this example, the QP in \eqref{eq:quadratic_program} has no inequality constraints and can therefore the KKT conditions can be directly solved.

\begin{figure}[t!]
	\hspace*{2cm}
	\includegraphics[width=0.5\columnwidth]{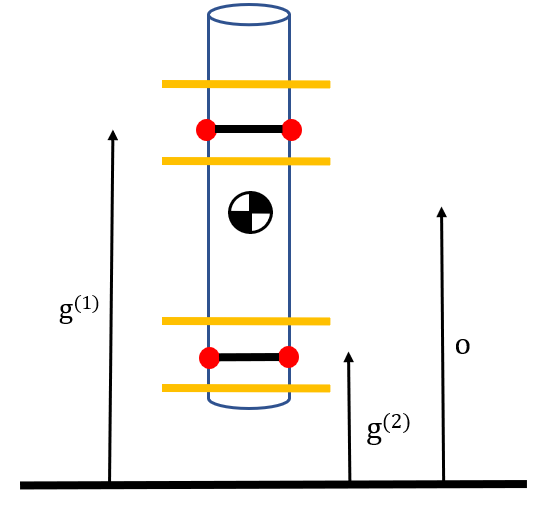}
	\caption{Lifting an object using two grippers indicated by red circles. The soft limits where the grippers should not cross are indicated by yellow lines.}
	\label{finger_picture}
\end{figure}

\begin{figure}[b!]
	\hspace*{0.3cm}
	\includegraphics[width=0.9\columnwidth]{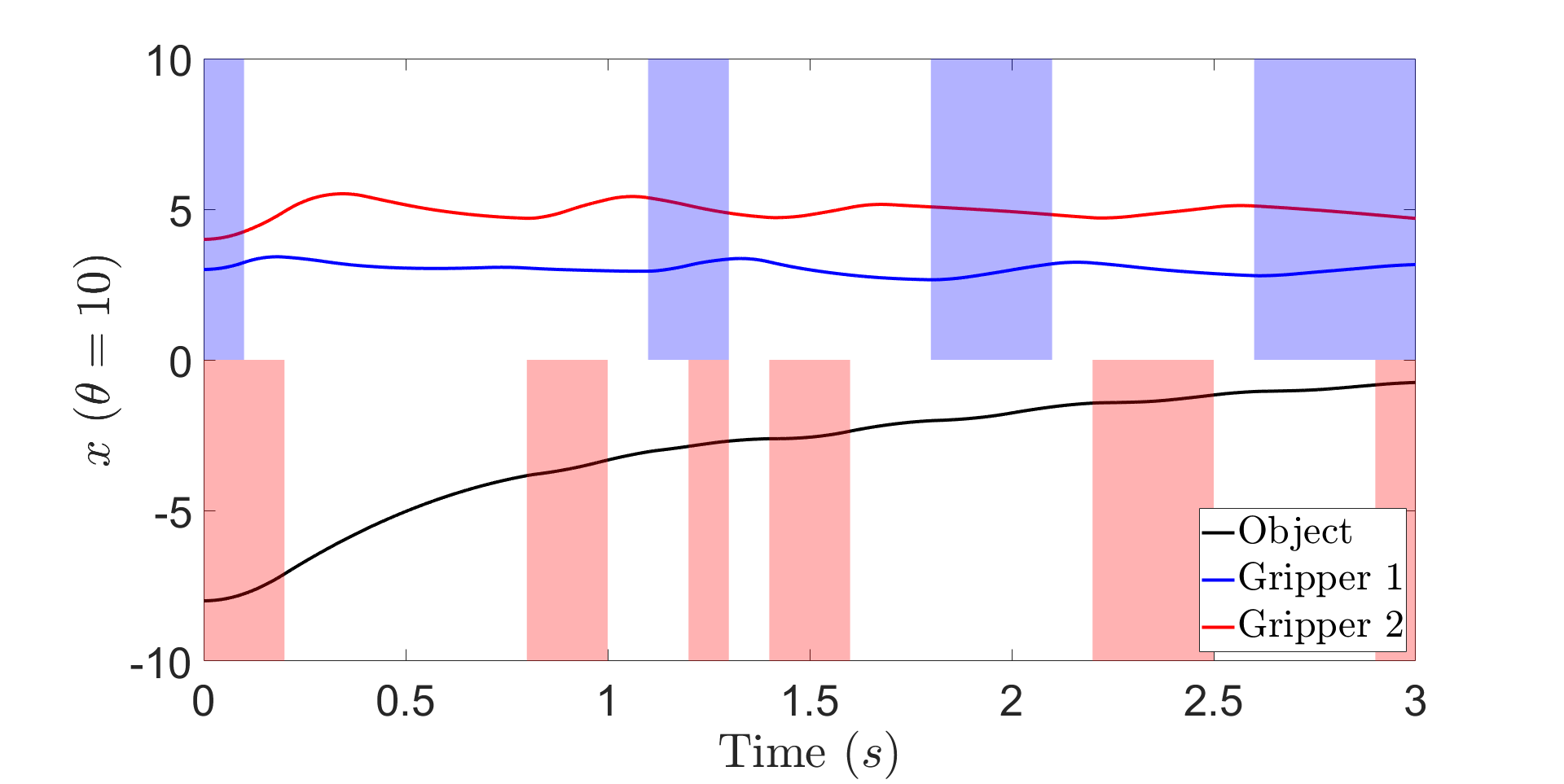}
	\caption{Finger gaiting with $\theta = 10$. Blue shading implies that gripper 1 is applying normal force to the object whereas red shading implies that gripper 2 is applying normal force.}
	\label{finger_1}
\end{figure}

\subsection{Finger Gaiting}

Next, we lift a rigid object upwards using four fingers. The setup for this problem is illustrated in Figure \ref{finger_picture}. The red circles indicate where the grippers interact the object and we assume that the grippers are always near the surface of the object and the force they apply on the object can be controlled. This force affects the friction between the object and grippers. Since the grippers never leave the surface, we assume that there is no rotation. 
The goal of this task is to lift the object vertically, while the fingers are constrained to stay close to their original locations (soft constraints shown in yellow). This task, therefore, requires finger gaiting to achieve large vertical motion of the object.

We use the formulation in \cite{stewart2000implicit} for modeling the system and denote the positions of the grippers as $g^{(1)}$, $g^{(2)}$ respectively and position of the object as $o$. We choose $g=9.81$ as gravitational acceleration and $\mu=1$ is the coefficient of friction for both grippers. 

\begin{figure}[b!]
	\hspace*{0.5cm}
	\includegraphics[width=0.8\columnwidth]{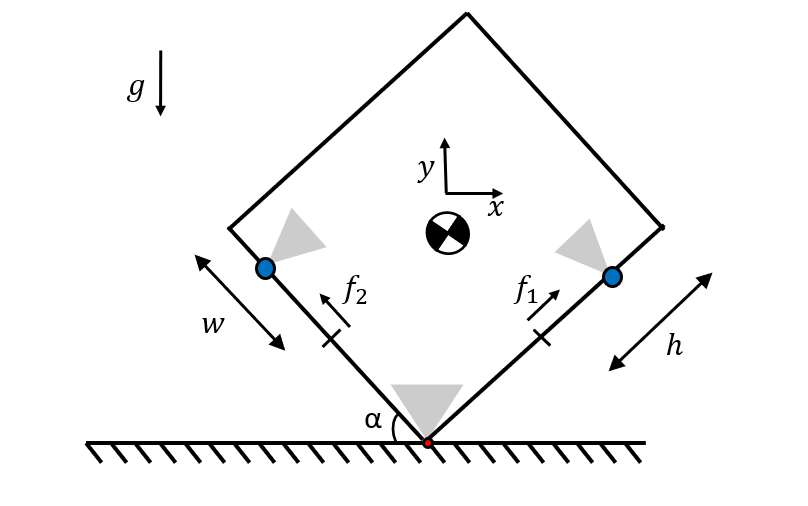}
	\caption{Pivoting a rigid object with two fingers (blue). The object can make and break contact with the ground and gray areas represent the friction cones.}
	\label{pivoting}
\end{figure}

We design a two different controllers based on Algorithm \ref{algortihm_ADMM} where $G = I$, $N = 10$. Both controllers use the MIQP projection method since other two projection methods failed to find stable motions.
For this example, we use $\theta = 10$, $\rho_k = 1.2$.  \rev{Parallelization leads to $\approx 2.5$x speedup for this particular example and we are able to run at $10$ Hz.} We also enforce limits:
\begin{align*}
	& 1 \leq g^{(1)} \leq 3, \; \forall k, \\
	& 3 \leq g^{(2)} \leq 5, \; \forall k.	
\end{align*}

We performed 100 trials starting from different initial conditions where $o(0)  \sim U[-6, -8]$, $g^{(1)}(0) \sim U[2, 3]$ , $g^{(2)}(0) \sim U[3, 4]$ are uniformly distributed and both the grippers and the object have zero initial speed. The controller managed to stabilize the system in all cases. As seen in Figure \ref{finger_1}, the grippers first throw the object into the air and then catch it followed by some finger gaiting.

\subsection{Pivoting}

In this example, we consider pivoting a rigid object that can make and break contact with the ground inspired by Hogan et. al \cite{hogan2020tactile}. Two fingers (indicated via blue) interact with the object as in Figure \ref{pivoting}. The goal is to balance the rigid-object at the midpoint.

The positions of the fingers with respect to the object are described via $f_1$, $f_2$ respectively. The normal force that the fingers exert onto the box can be controlled. The center of mass position is denoted by $x$ and $y$ respectively, $\alpha$ denotes the angle with the ground and $w=1$, $h=1$ are the dimensions of the object. The coefficient of friction for the fingers are $\mu_1 = \mu_2 = 0.1$, and the coefficient of friction with the ground is $\mu_3 = 1$. We take the gravitational acceleration as $g=9.81$ and mass of the object as $m=1$. The fingers start close to the pivot point where $f_1 = -0.3$, $f_2 = -0.7$ and the objects configuration is given by $x = 0$, $y=1.36$ and $\alpha = 0.2$. The goal is to balance the object at the midpoint ($x=0$, $y = \sqrt{2}$,  $\alpha = \pi/4$) while simultaneously moving the fingers towards the end of the object $(f_1 = f_2 = 0.9)$.

\begin{figure}[t!]
	\hspace*{-0.5cm}
	\includegraphics[width=1.2\columnwidth]{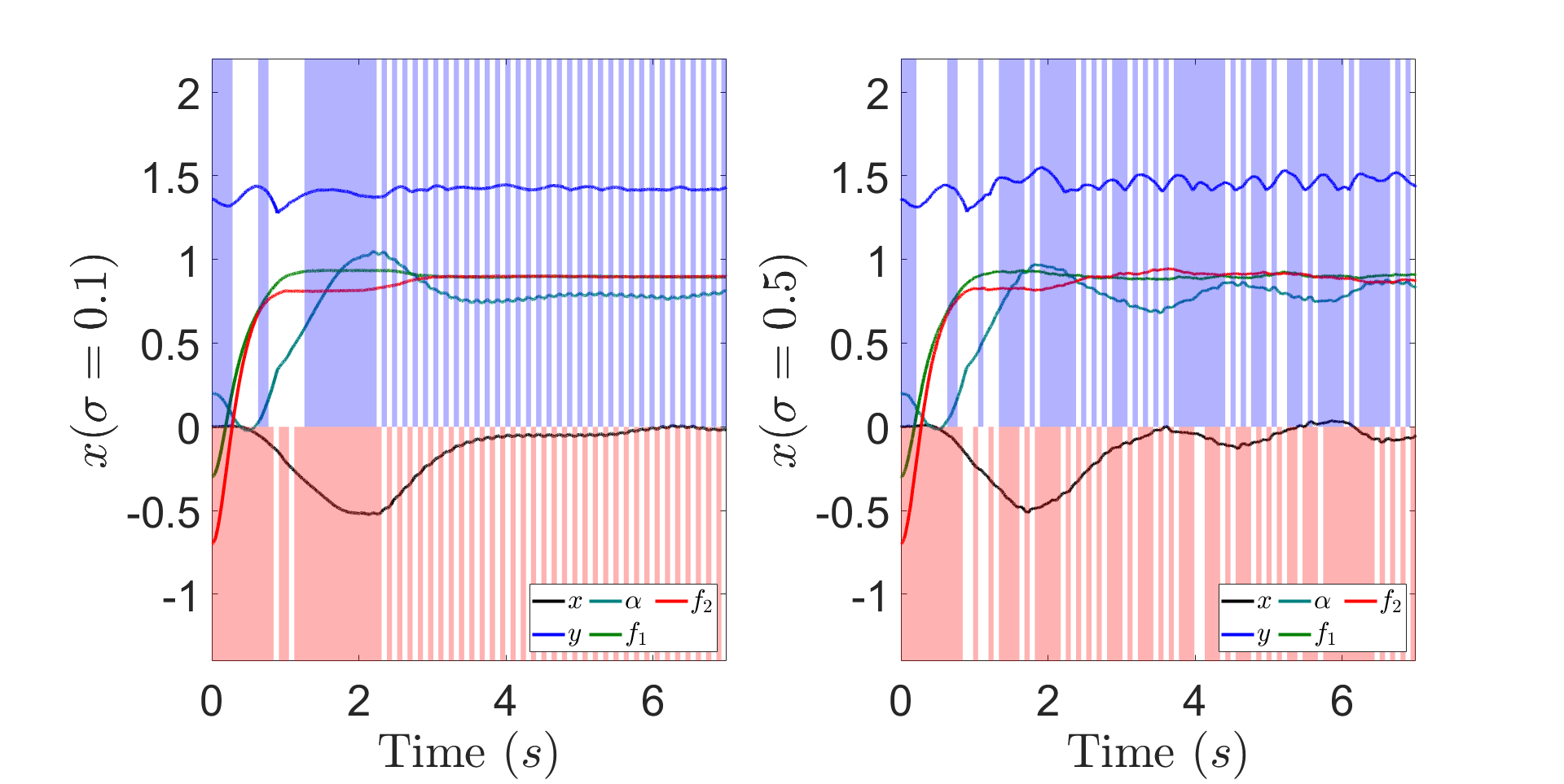}
	\caption{Pivoting example, Gaussian disturbances with standard deviations $\sigma$. Blue shading implies that gripper 1 is applying normal force to the object whereas red shading implies that gripper 2 is applying normal force.}
	\label{pivoting_random}
\end{figure}

We model the system using an implicit time-stepping scheme \cite{stewart2000implicit}. The system has 3 contacts, where the finger contacts are represented by $3$ complementarity variables each, and the ground contact is modeled via $4$ complementarity variables. The system has ten states ($n_x = 10$), 10 complementarity variables ($n_\lambda = 10$) and 4 inputs ($n_u = 4$). More concretely, it has $36$ hybrid modes.

For this example, as the LCS-representation is only an approximation, we compute a new local LCS approximation at every time step $k$. We pick $\theta = 5$, $N = 10$, $\rho_k = 1.1$ and use the local LCS approximation given at time-step $k$ while planning. \rev{Parallelization leads to $\approx 4$x speedup for this particular example and controller can run around $16$ Hz.}

Figure \ref{pivoting_random} demonstrates the robustness of the controller for different Gaussian disturbances (added to dynamics) with standard deviations (for $\sigma=0.1,0.5$). Note that at every time step, all positions and velocities (including angular) are affected by the process noise. The object approximately reaches the desired configuration (midpoint where $\alpha=\frac{\pi}{4}$) for $\sigma=0,0.05,0.1$ and starts failing to get close to the desired configuration for $\sigma=0.5$. Plots with $\sigma=0$ and $\sigma = 0.05$ are omitted as those were similar to the one with $\sigma = 0.1$. We emphasize that the process noise causes unplanned mode changes and the controller seamlessly reacts.
This example demonstrates that our method works well with successive linearizations as many multi-contact systems can not be captured via a single LCS approximation.

\section{Experimental Validation}

\begin{figure}[t!]
	\hspace*{1.3cm}
	\includegraphics[width=0.7\columnwidth]{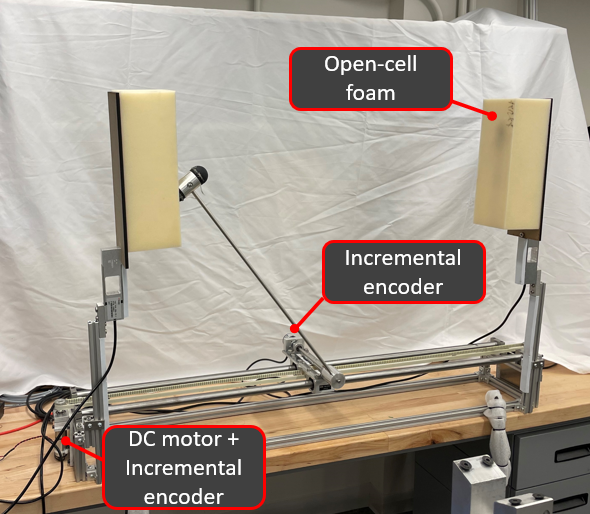}
	\caption{Experimental setup for cart-pole with soft walls.}
	\label{cartpole_experiment}
\end{figure}

We test the multi-contact MPC algorithm on an experimental cart-pole system with soft walls shown in Figure \ref{cartpole_experiment} replicating the system described in \ref{subsec:cartpole_sim}. A DC motor with a belt drive generates the linear motion of the cart. Soft walls are made of open-cell polyurethane foam.

\begin{figure}[b!]
	\includegraphics[width=1\columnwidth]{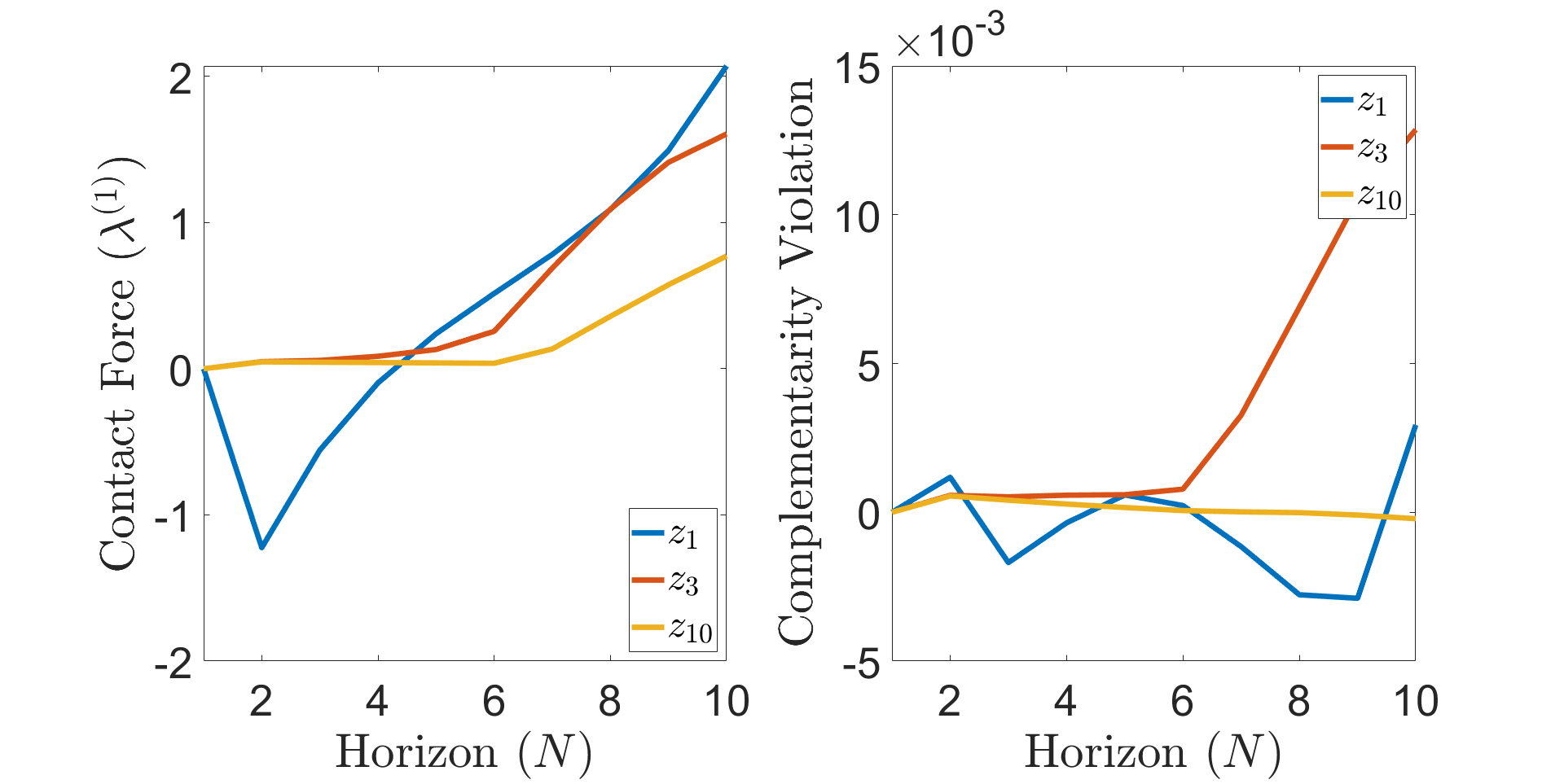}
	\caption{Evolution of contact force and complementarity violation during ADMM iteration when the cart is close to a contact surface. }
	\label{mm}
\end{figure}

We consider the linear model as in \ref{subsec:cartpole_sim} (also used in \cite{aydinoglu2020stabilization}) where $m_c = 0.978$ is the mass of the cart, $m_p = 0.411$ is the mass of the pole and the rod, $l_p = 0.6$ is the length of the pole, $l_c = 0.4267$ is the length of the center of mass position, $k_1 = k_2 = 100$ are the stiffness parameters of the walls and $d = 0.39$ is the distance between the origin and walls.

The parameters of the MPC algorithm are $N = 10$, $\rho = 2.3$, $\theta = 10$, and $G = 0.5 I$. We use the LCP-based projection method, and solve the quadratic programs via utilizing the KKT system. We note that the algorithm is capable of running faster than $240$ Hz as mentioned previously, but due to high communication delays between our motor controller and computer, we run the system at $100$ Hz. In Figure \ref{mm}, we illustrate, for one particular state, the evolution of the contact force throughout the ADMM process (at ADMM steps $1$, $3$, and $10$).
Notice that as the the algorithm progresses, contact forces become more realistic.

We initialize the cart at the origin and introduce random perturbations to cover a wide range of initial conditions that lead to contact events. Specifically, we start the cart-pole at the origin where our controller is active. Then, we apply an input disturbance $u_d \sim U [10, 15]$ for $250$ ms to force contact events. We repeated this experiment $10$ times and our controller managed to stabilize the system in all trials. 

To empirically evaluate the gap between C3, which is sub-optimal, and true solutions to \eqref{eq:MPC_original}, and to assess the impact of modeling errors, we report the cost-to-go values for our method, MIQP solution (as in \eqref{eq:MPC_MIQP}), and the actual observed states. 
More precisely, given the current state of the nonlinear plant, we calculate the cost as in \eqref{eq:MPC_original} using the inputs recovered from both the C3 algorithm and MIQP algorithm. Since C3 algorithm is not guaranteed to produce a $(x,u,\lambda)$ that strictly satisfies complementarity, the predicted cost may not match the simulated cost once $u$ is applied.
For the actual plant, we use the data from $N$ steps into the future and calculate the same cost. 
Notice that even though the cost-to-go of MIQP solution is always lower, as expected, the optimality gap is fairly small. 
This highlights that C3 finds near-optimal solutions, at least for this particular problem.
Also, notice that the actual cost-to-go matches the planned one which further motivates the applicability of LCS representations in model-based control for nonlinear multi-contact systems.

\begin{figure}[t!]
	\includegraphics[width=1\columnwidth]{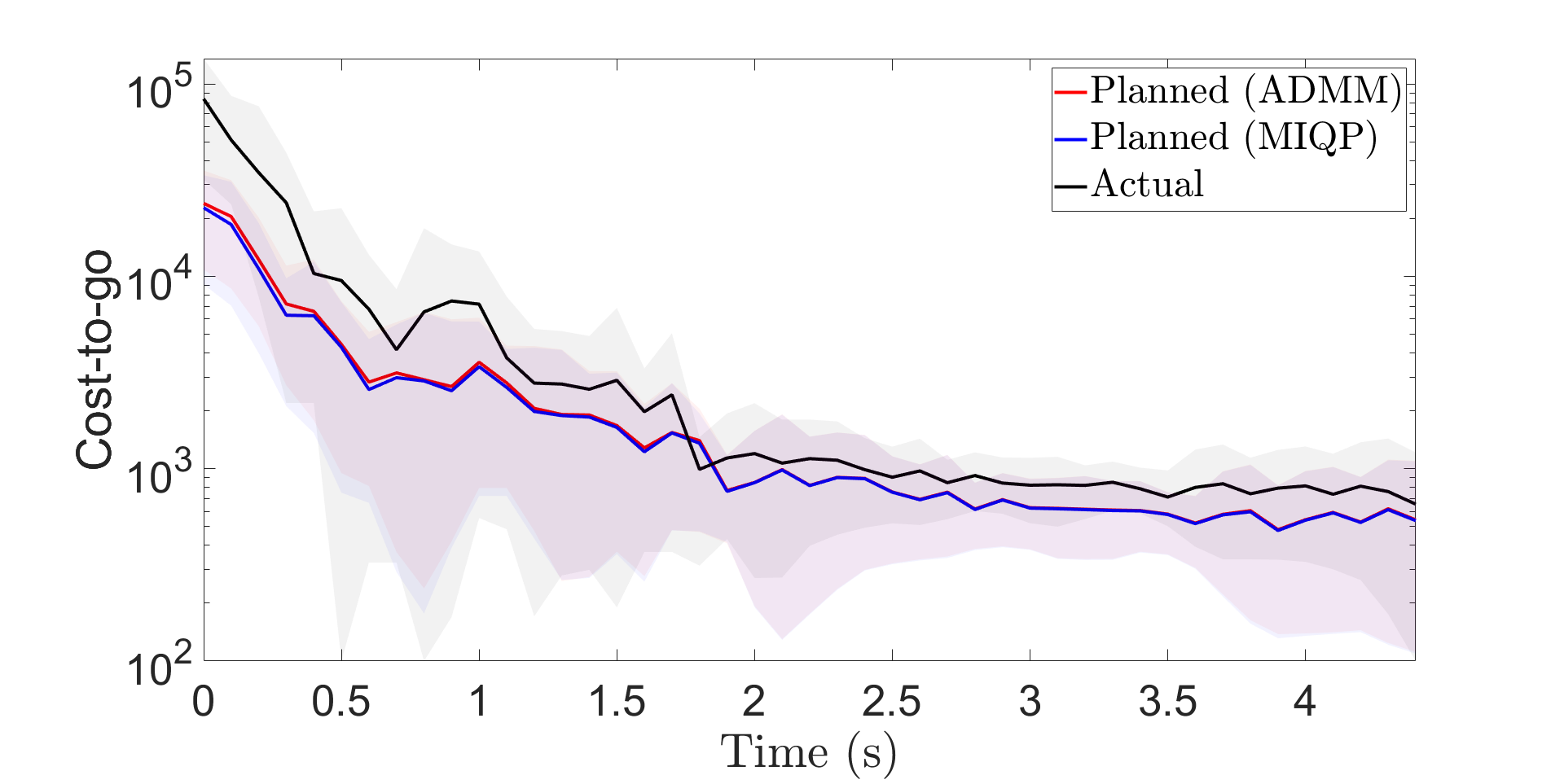}
	\caption{Cost-to-go values for the cart-pole experiment.}
	\label{cartpole_cost_to_go}
\end{figure}

\section{Conclusion}

In this work, we present an algorithm, C3, for model predictive control of multi-contact systems. The algorithm relies on solving QP's accompanied by projections and both can be solved efficiently for multi-contact systems. The effectiveness of our approach is verified on three numerical examples and our results are validated on an experimental setup. For fairly complex examples with frictional contact, our method has a fast run-time. We also demonstrate with the experiment that our heuristic can find close-to-optimal strategies.

The framework tackles the hybrid MPC problem by shifting the complexity to the projection sub-problems. While projections are decoupled in time, they are still difficult to solve. Here, we demonstrate three potential approaches to this projection stage, but exploring alternative heuristics is of future interest.
In addition, the choice of parameters (such as $G$ and $U$) greatly affects the performance of the algorithm and exploring different approaches for deciding on parameters is of future interest too.
Further evaluation, for instance on true dexterous manipulation, and the integration with learned models \cite{pfrommer2020contactnets} is in the scope of future work. Additionally, we anticipate that code optimization will lead significant improvements (3x or more) in control rate. 

\section*{Acknowledgements}

We thank Philip Sieg for his help with the cart-pole experimental setup, Tianze Wang and Yike Li for their earlier explorations of ADMM-based multi-contact optimal control, Mathew Halm and William Yang for helpful discussions related to the pivoting example and visualization code.


\addtolength{\textheight}{-1cm}  

\bibliographystyle{ieeetr}
\bibliography{Refs}

\begin{thebibliography}{10}

\bibitem{posa2014direct}
M.~Posa, C.~Cantu, and R.~Tedrake, ``A direct method for trajectory
  optimization of rigid bodies through contact,'' {\em The International
  Journal of Robotics Research}, vol.~33, no.~1, pp.~69--81, 2014.

\bibitem{garcia1989model}
C.~E. Garcia, D.~M. Prett, and M.~Morari, ``Model predictive control: Theory
  and practice—a survey,'' {\em Automatica}, vol.~25, no.~3, pp.~335--348,
  1989.

\bibitem{borrelli2017predictive}
F.~Borrelli, A.~Bemporad, and M.~Morari, {\em Predictive control for linear and
  hybrid systems}.
\newblock Cambridge University Press, 2017.

\bibitem{ding2019real}
Y.~Ding, A.~Pandala, and H.-W. Park, ``Real-time model predictive control for
  versatile dynamic motions in quadrupedal robots,'' in {\em 2019 International
  Conference on Robotics and Automation (ICRA)}, pp.~8484--8490, IEEE, 2019.

\bibitem{zhakatayev2017successive}
A.~Zhakatayev, B.~Rakhim, O.~Adiyatov, A.~Baimyshev, and H.~A. Varol,
  ``Successive linearization based model predictive control of variable
  stiffness actuated robots,'' in {\em 2017 IEEE international conference on
  advanced intelligent mechatronics (AIM)}, pp.~1774--1779, IEEE, 2017.

\bibitem{mastalli2020crocoddyl}
C.~Mastalli, R.~Budhiraja, W.~Merkt, G.~Saurel, B.~Hammoud, M.~Naveau,
  J.~Carpentier, L.~Righetti, S.~Vijayakumar, and N.~Mansard, ``Crocoddyl: An
  efficient and versatile framework for multi-contact optimal control,'' in
  {\em 2020 IEEE International Conference on Robotics and Automation (ICRA)},
  pp.~2536--2542, IEEE, 2020.

\bibitem{sleiman2021unified}
J.-P. Sleiman, F.~Farshidian, M.~V. Minniti, and M.~Hutter, ``A unified mpc
  framework for whole-body dynamic locomotion and manipulation,'' {\em IEEE
  Robotics and Automation Letters}, vol.~6, no.~3, pp.~4688--4695, 2021.

\bibitem{winkler2018gait}
A.~W. Winkler, C.~D. Bellicoso, M.~Hutter, and J.~Buchli, ``Gait and trajectory
  optimization for legged systems through phase-based end-effector
  parameterization,'' {\em IEEE Robotics and Automation Letters}, vol.~3,
  no.~3, pp.~1560--1567, 2018.

\bibitem{bledt2020regularized}
G.~Bledt, {\em Regularized predictive control framework for robust dynamic
  legged locomotion}.
\newblock PhD thesis, Massachusetts Institute of Technology, 2020.

\bibitem{lee2020learning}
J.~Lee, J.~Hwangbo, L.~Wellhausen, V.~Koltun, and M.~Hutter, ``Learning
  quadrupedal locomotion over challenging terrain,'' {\em Science robotics},
  vol.~5, no.~47, 2020.

\bibitem{kuindersma2016optimization}
S.~Kuindersma, R.~Deits, M.~Fallon, A.~Valenzuela, H.~Dai, F.~Permenter,
  T.~Koolen, P.~Marion, and R.~Tedrake, ``Optimization-based locomotion
  planning, estimation, and control design for the atlas humanoid robot,'' {\em
  Autonomous robots}, vol.~40, no.~3, pp.~429--455, 2016.

\bibitem{hogan2020reactive}
F.~R. Hogan and A.~Rodriguez, ``Reactive planar non-prehensile manipulation
  with hybrid model predictive control,'' {\em The International Journal of
  Robotics Research}, vol.~39, no.~7, pp.~755--773, 2020.

\bibitem{marcucci2017approximate}
T.~Marcucci, R.~Deits, M.~Gabiccini, A.~Bicchi, and R.~Tedrake, ``Approximate
  hybrid model predictive control for multi-contact push recovery in complex
  environments,'' in {\em 2017 IEEE-RAS 17th International Conference on
  Humanoid Robotics (Humanoids)}, pp.~31--38, IEEE, 2017.

\bibitem{aydinoglu2021stability}
A.~Aydinoglu, M.~Fazlyab, M.~Morari, and M.~Posa, ``Stability analysis of
  complementarity systems with neural network controllers,'' in {\em
  Proceedings of the 24th International Conference on Hybrid Systems:
  Computation and Control}, pp.~1--10, 2021.

\bibitem{tassa2012synthesis}
Y.~Tassa, T.~Erez, and E.~Todorov, ``Synthesis and stabilization of complex
  behaviors through online trajectory optimization,'' in {\em 2012 IEEE/RSJ
  International Conference on Intelligent Robots and Systems}, pp.~4906--4913,
  IEEE, 2012.

\bibitem{koenemann2015whole}
J.~Koenemann, A.~Del~Prete, Y.~Tassa, E.~Todorov, O.~Stasse, M.~Bennewitz, and
  N.~Mansard, ``Whole-body model-predictive control applied to the hrp-2
  humanoid,'' in {\em 2015 IEEE/RSJ International Conference on Intelligent
  Robots and Systems (IROS)}, pp.~3346--3351, IEEE, 2015.

\bibitem{kumar2014real}
V.~Kumar, Y.~Tassa, T.~Erez, and E.~Todorov, ``Real-time behaviour synthesis
  for dynamic hand-manipulation,'' in {\em 2014 IEEE International Conference
  on Robotics and Automation (ICRA)}, pp.~6808--6815, IEEE, 2014.

\bibitem{marcucci2020warm}
T.~Marcucci and R.~Tedrake, ``Warm start of mixed-integer programs for model
  predictive control of hybrid systems,'' {\em IEEE Transactions on Automatic
  Control}, 2020.

\bibitem{cleac2021linear}
S.~L. Cleac'h, T.~Howell, M.~Schwager, and Z.~Manchester, ``Linear
  contact-implicit model-predictive control,'' {\em arXiv preprint
  arXiv:2107.05616}, 2021.

\bibitem{frick2019low}
D.~Frick, A.~Georghiou, J.~L. Jerez, A.~Domahidi, and M.~Morari,
  ``Low-complexity method for hybrid mpc with local guarantees,'' {\em SIAM
  Journal on Control and Optimization}, vol.~57, no.~4, pp.~2328--2361, 2019.

\bibitem{boyd2011distributed}
S.~Boyd, N.~Parikh, and E.~Chu, {\em Distributed optimization and statistical
  learning via the alternating direction method of multipliers}.
\newblock Now Publishers Inc, 2011.

\bibitem{stellato2018embedded}
B.~Stellato, V.~V. Naik, A.~Bemporad, P.~Goulart, and S.~Boyd, ``Embedded
  mixed-integer quadratic optimization using the osqp solver,'' in {\em 2018
  European Control Conference (ECC)}, pp.~1536--1541, IEEE, 2018.

\bibitem{takapoui2020simple}
R.~Takapoui, N.~Moehle, S.~Boyd, and A.~Bemporad, ``A simple effective
  heuristic for embedded mixed-integer quadratic programming,'' {\em
  International journal of control}, vol.~93, no.~1, pp.~2--12, 2020.

\bibitem{park2017general}
J.~Park and S.~Boyd, ``General heuristics for nonconvex quadratically
  constrained quadratic programming,'' {\em arXiv preprint arXiv:1703.07870},
  2017.

\bibitem{heemels2000linear}
W.~Heemels, J.~M. Schumacher, and S.~Weiland, ``Linear complementarity
  systems,'' {\em SIAM journal on applied mathematics}, vol.~60, no.~4,
  pp.~1234--1269, 2000.

\bibitem{stewart2000implicit}
D.~Stewart and J.~C. Trinkle, ``An implicit time-stepping scheme for rigid body
  dynamics with coulomb friction,'' in {\em Proceedings 2000 ICRA. Millennium
  Conference. IEEE International Conference on Robotics and Automation.
  Symposia Proceedings (Cat. No. 00CH37065)}, vol.~1, pp.~162--169, IEEE, 2000.

\bibitem{pfrommer2020contactnets}
S.~Pfrommer, M.~Halm, and M.~Posa, ``Contactnets: Learning discontinuous
  contact dynamics with smooth, implicit representations,'' {\em arXiv preprint
  arXiv:2009.11193}, 2020.

\bibitem{cottle2009linear}
R.~W. Cottle, J.-S. Pang, and R.~E. Stone, {\em The linear complementarity
  problem}.
\newblock SIAM, 2009.

\bibitem{halm2019modeling}
M.~Halm and M.~Posa, ``Modeling and analysis of non-unique behaviors in
  multiple frictional impacts,'' {\em arXiv preprint arXiv:1902.01462}, 2019.

\bibitem{cauligi2020learning}
A.~Cauligi, P.~Culbertson, B.~Stellato, D.~Bertsimas, M.~Schwager, and
  M.~Pavone, ``Learning mixed-integer convex optimization strategies for robot
  planning and control,'' in {\em 2020 59th IEEE Conference on Decision and
  Control (CDC)}, pp.~1698--1705, IEEE, 2020.

\bibitem{diamond2018general}
S.~Diamond, R.~Takapoui, and S.~Boyd, ``A general system for heuristic
  minimization of convex functions over non-convex sets,'' {\em Optimization
  Methods and Software}, vol.~33, no.~1, pp.~165--193, 2018.

\bibitem{osqp}
B.~Stellato, G.~Banjac, P.~Goulart, A.~Bemporad, and S.~Boyd, ``{OSQP}: an
  operator splitting solver for quadratic programs,'' {\em Mathematical
  Programming Computation}, vol.~12, no.~4, pp.~637--672, 2020.

\bibitem{Gurobi}
{Gurobi Optimization, LLC}, ``{Gurobi Optimizer Reference Manual},'' 2021.

\bibitem{dirkse1995path}
S.~P. Dirkse and M.~C. Ferris, ``The path solver: a nommonotone stabilization
  scheme for mixed complementarity problems,'' {\em Optimization methods and
  software}, vol.~5, no.~2, pp.~123--156, 1995.

\bibitem{aydinoglu2020stabilization}
A.~Aydinoglu, P.~Sieg, V.~M. Preciado, and M.~Posa, ``Stabilization of
  complementarity systems via contact-aware controllers,'' {\em arXiv preprint
  arXiv:2008.02104}, 2020.

\bibitem{li2021model}
H.~Li, R.~J. Frei, and P.~M. Wensing, ``Model hierarchy predictive control of
  robotic systems,'' {\em IEEE Robotics and Automation Letters}, vol.~6, no.~2,
  pp.~3373--3380, 2021.

\bibitem{hogan2020tactile}
F.~R. Hogan, J.~Ballester, S.~Dong, and A.~Rodriguez, ``Tactile dexterity:
  Manipulation primitives with tactile feedback,'' in {\em 2020 IEEE
  international conference on robotics and automation (ICRA)}, pp.~8863--8869,
  IEEE, 2020.

\end{thebibliography}

\end{document}